\documentclass[letterpaper, 10 pt, conference]{ieeeconf}  

\IEEEoverridecommandlockouts                              

\usepackage{graphics} 
\usepackage{epsfig} 
\usepackage{mathptmx} 
\usepackage{times} 
\usepackage{amsmath} 
\usepackage{amssymb}  
\usepackage{algorithm}
\usepackage{algorithmic}
\usepackage{newtxtext, newtxmath}
\usepackage{caption}
\usepackage{subfigure}

\newtheorem{theorem}{Theorem}[section]
\newtheorem{definition}[theorem]{Definition}

\title{\LARGE \bf
Learning Task Specifications from Demonstrations as Probabilistic Automata
}

\author{Mattijs Baert, Sam Leroux, Pieter Simoens
\thanks{All authors are with IDLab, Department of Information Technology at Ghent University - imec,
        Technologiepark 126, B-9052 Ghent, Belgium
        {\tt\small mattijs.baert@ugent.be}}%
}

\begin{document}

\maketitle
\thispagestyle{empty}
\pagestyle{empty}

\begin{abstract}
Specifying tasks for robotic systems traditionally requires coding expertise, deep domain knowledge, and significant time investment. While learning from demonstration offers a promising alternative, existing methods often struggle with tasks of longer horizons. To address this limitation, we introduce a computationally efficient approach for learning probabilistic deterministic finite automata (PDFA) that capture task structures and expert preferences directly from demonstrations. Our approach infers sub-goals and their temporal dependencies, producing an interpretable task specification that domain experts can easily understand and adjust. We validate our method through experiments involving object manipulation tasks, showcasing how our method enables a robot arm to effectively replicate diverse expert strategies while adapting to changing conditions.
\end{abstract}

\section{INTRODUCTION}
\begin{figure*}
    \centering
    \includegraphics[width=0.9\linewidth]{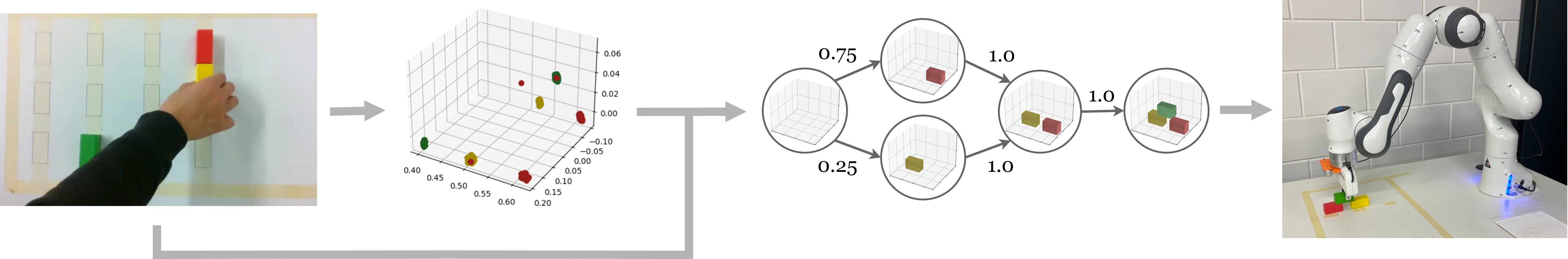}
    \vspace{-0.05cm}
    \caption{1) Given a set of demonstrations, potential sub-goals are extracted through clustering. 2) From the set of sub-goals and the demonstrations a PDFA is constructed representing task structure and demonstrator preferences. 3) The learned PDFA can be used for task planning.}
    \label{fig:method}
\vspace{-0.6cm}
\end{figure*}
Robots have become indispensable in manufacturing assembly tasks, yet they often operate under rigidly fixed sequences of sub-tasks. This rigidity causes inefficiencies, such as robots idling when a tool or sub-assembly is unavailable, despite other tasks being ready for execution. For example, in a scenario where two sub-assemblies must be constructed separately and then combined into a final product, a robot constrained by a fixed task sequence would be forced to wait if materials for the first sub-assembly are unavailable, even though it could begin work on the second sub-assembly. Encoding these temporal dependencies into a task specification presents significant challenges for domain experts. Additionally, the variability in execution sequences by different operators renders traditional Learning from Demonstration (LfD) techniques inadequate \cite{ho2016generative}.
\\
To address these challenges, we propose a method for identifying sub-goals and their temporal dependencies from demonstrations. Specifically, we present an algorithm to construct a Probabilistic Deterministic Finite Automaton (PDFA) that models both the task structure and the demonstrators' preferences. In this model, state transitions represent the completion of sub-goals, while states denote the sub-goals that have already been completed. Each transition is assigned a probability reflecting the demonstrators' preferences. This PDFA can then be used to plan and replicate expert behaviors. Utilizing a PDFA as a task representation offers several advantages. Firstly, it provides an interpretable model that domain experts can easily validate and modify. Secondly, PDFAs facilitate online planning, enabling robots to adapt their behavior based on changing conditions. Lastly, our method accommodates diversity in demonstrations, enhancing the flexibility and robustness of autonomous robots. 
\\
Consider the task of building a tower with three colored blocks, where the green block has to be placed on top and the red and yellow blocks form the base (as shown in Fig. \ref{fig:method}). This task can be completed in various ways; for instance, some individuals may start by placing the red block first, while others may begin with the yellow block.
To learn a specification for this task, we start by identifying a set of sub-goals, which, in this case, involve placing each block in its correct position. We hypothesize that genuine sub-goals correspond to partial states frequently observed in expert demonstrations, as supported by prior research \cite{ghazanfari2020sequential}. To identify these sub-goals, we apply clustering techniques to group similar states and identify the most representative ones.
We then construct a PDFA that captures the temporal dependencies between these sub-goals. The variability observed in the demonstrations is reflected in the PDFA's structure, allowing the robot to choose which block to place first while ensuring the task always concludes with the green block on top.
The transition probabilities in the PDFA reflect the preference of the demonstrator, indicating preference of placing the red block before the yellow block.
The inferred PDFA can then be used by an autonomous agent, such as a robot arm, to execute the learned task.
\\
Our method addresses the challenge of learning compositional long-horizon tasks, a capability often beyond traditional LfD approaches \cite{ho2016generative,DBLP:conf/icra/ZhangMJLCGA18,kober2009learning}. Temporal logics provide a framework for representing such complex task specifications. However, existing methods for inferring temporal logic formulas from demonstrations generally require extensive knowledge of the robot's dynamics \cite{chou2022learning, liu2024interpretable}, are limited to simpler tasks \cite{jha2019telex}, or necessitate manual specification of sub-goals \cite{shah2018bayesian, roy2023learning}. While temporal logics are generally translated into automata for task planning \cite{baier2008principles}, our approach directly infers an automaton. Traditional methods for Deterministic Finite Automaton (DFA) inference can be computationally intensive \cite{lang1998results,petasis2004eg,watanabe2021probabilistic}. By assuming that the extracted sub-goals represent the set of valid transitions, we improve both efficiency and scalability of inferring DFA task specifications. 
\\
The main contribution of this paper is the development of a method for learning sub-goals and their temporal dependencies from demonstrations. This involves the construction of a PDFA that models task structure and demonstrator preferences. This approach accommodates variability in task execution, enhancing flexibility and robustness in robot task planning. Additionally, by introducing an inductive bias based on sub-goals, the method improves the efficiency and scalability of automaton inference for LfD. We demonstrate the practical application of this approach on a variety of object manipulation tasks, showcasing its effectiveness in real-world scenarios.

\section{RELATED WORK}
\label{sec:related}
Many LfD studies focus on learning a policy that maps agent states to actions using techniques such as Reinforcement Learning (RL) \cite{DBLP:conf/icra/ZhangMJLCGA18,DBLP:conf/rss/MandlekarXMS020} or movement primitives \cite{kober2009learning,schaal2006dynamic}. Another set of LfD methods frames the problem as acquiring a standard Markovian reward function \cite{ho2016generative,abbeel2004apprenticeship,baert2023maximum}, which can then be used to learn a policy. However, extracting task specifications from a learned reward function or policy is non-trivial \cite{arnold2017value}, and these methods are generally limited to tasks of limited duration.
\\
To broaden the scope of LfD and tackle more complex tasks, significant efforts have been made to explore more expressive models, such as temporal logics \cite{pnueli1977temporal}, for defining task specifications. Temporal logics have been used for goal definitions in domain-independent planning \cite{kim2017collaborative}, synthesis of reactive systems \cite{kress2009temporal}, and RL \cite{aksaray2016q,li2017reinforcement,xiong2022constrained}. However, writing correct formal specifications requires domain knowledge. Recent studies have proposed learning Linear Temporal Logic (LTL) formulas from data, often relying on labeled datasets of positive and negative traces \cite{bombara2021offline,kong2016temporal,yan2022neuro}. In the context of LfD, collecting negative examples can be challenging and dangerous.
Recent methods have focused on learning LTL exclusively from positive examples \cite{shah2018bayesian,roy2023learning,vazquez2018learning}. However, these approaches can be slow due to the extensive formula exploration space or limit their search to a restricted set of formula templates. Additionally, these methods assume a world state description in terms of Boolean propositions for each time step in the demonstrations. Studies like \cite{jha2019telex} and \cite{chiu2023temporal} allow learning Signal Temporal Logic (STL) formulas from continuous, unstructured demonstrations but are limited to very simple formulas. The method of \cite{chou2022learning} can learn both the formula structure and the set of propositions mapped to the world state but requires complete knowledge of the environment's dynamics. In contrast, our method only makes certain assumptions about the task structure and underlying sub-goals.
\\
To use temporal logic formulas for planning, they typically need to be translated to automata, which can lead to the state-explosion problem \cite{baier2008principles}. One way to overcome this is by directly learning a DFA. \cite{araki2019learning} proposes a method for learning transitions between states given the DFA structure. \cite{watanabe2021probabilistic} extends DFA inference based on state merging to learn PDFAs under safety constraints. However, the merging process can be computationally expensive and may lead to exponential growth in complexity with the number of states.
\\
Similar to sub-goal inference in our method, many methods for learning plan specifications from demonstrations first segment each demonstration into sub-tasks. Previous work relies on manual segmentation \cite{ekvall2008robot}, hidden Markov models \cite{niekum2015learning,konidaris2012robot}, Chinese restaurant processes \cite{grollman2010incremental}, or seq-2-seq models \cite{pirk2020modeling}. \cite{manschitz2014learning} uses support vector machines and Gaussian mixture models to assign each time step to a motion primitive from a movement library, while \cite{mohseni2019simultaneous} uses narrations of the human demonstrator to identify boundaries between primitives. Unlike our method, these techniques require extensive hyperparameter tuning, consider only sequential task structures, or demand additional efforts from the demonstrator.
\\
There is extensive research on inferring DFAs from examples, which extends beyond the scope of LfD. As demonstrations typically consist only of valid trajectories, we focus on methods for learning DFAs from positive examples. Traditional DFA inference methods include state-merging algorithms that iteratively generalize a DFA by merging states \cite{lang1992random,lang1998results}, incremental approaches that refine a set of candidate automata as new examples are introduced \cite{rivest1989inference}, and genetic programming techniques that evolve DFAs over time \cite{dupont1994regular,petasis2004eg}. These methods are versatile, making few assumptions about the underlying system generating the traces, and thus are applicable across various domains. However, this generality often comes at the cost of significant computational resources and time. Our approach aims to improve the efficiency and scalability of DFA inference in the context of LfD by introducing specific assumptions about the task structure and sub-goals.

\section{PROBLEM FORMULATION}
\label{sec:problem}
\subsection{States, Demonstrations and Sub-Goals}
Let $S$ be the set of world states. Each world state $s \in S$ is represented as a vector $s=(f_0,f_1,...,f_{n-1})$, where $f_i \in \mathcal{F}$ are features describing the state, and $\mathcal{F}$ denotes the set of all features. These features could include positions, object properties, agent statuses, environmental conditions, etc. 
A demonstration $d=\{s_0,s_1,...,s_{T-1}\}$ is defined as a series of states with $T$ denoting the duration of the demonstration measured in discrete time steps. A sub-goal $g=(c,r)$ is defined by a center point $c$ using a subset of features $\mathcal{F}_{g} \subseteq \mathcal{F}$ and a radius $r$, forming an $\mid \mathcal{F}_g \mid$-dimensional ball. We define a mapping function $\phi(\mathcal{F}_g, s)$ which extracts, from a state $s$, a partial state $\hat{s}$ consisting of the features specified by $\mathcal{F}_g$. 
To determine if a state $s$ satisfies a sub-goal, we check if the Euclidean distance between the partial state $\hat{s} = \phi(\mathcal{F}_g, s)$ and the center is smaller or equal to the specified radius $r$.

\subsection{Task Specification}
We assume that the human demonstrator has a task in mind that can be completed in finite time. Our objective is to learn this task in the form of a DFA, which captures the valid sequences of sub-goals as demonstrated by the experts.
\begin{definition}{(DFA)}
    A Deterministic Finite Automaton (DFA) is a tuple $\mathcal{A} = (Q, q_0, \Sigma, \delta, F)$, where $Q$ is a finite set of states, $q_0 \in Q$ is the initial state, $\Sigma$ is the input alphabet, $\delta : Q \times \Sigma \rightarrow Q$ is the transition function and $F \subseteq Q$ is the set of accepting states.
\end{definition}
A word $\omega$ is composed of a finite sequence of symbols $\sigma \in \Sigma$: $\omega = \sigma_0 \sigma_1...\sigma_{n-1}$. We define $\epsilon$ as the empty word. The concatenation of a symbol $\sigma$ to a word $\omega$ is defined as $\omega^{\prime} = \omega \cdot \sigma$. A word $\omega$ generates a trajectory of DFA states $\tau = q_0q_1...q_{n}$ if $q_{n} = \delta(q_{n-1}, \sigma_{n-1})$ for all $n \geq 0$. A finite input word $\omega$ is accepted by the automaton if the corresponding generated trajectory ends in a state within the set of accepting states: $q_{n} \in F$. The (accepted) language of $\mathcal{A}$, denoted by $\mathcal{L}(\mathcal{A})$, is the set of all accepted input words.
In addition to learning the DFA, we aim to encode the demonstrator's preferences by quantifying the probabilities of transitions between DFA states. Thus, our goal is to learn a PDFA that captures both the valid sequences (accepting words) and their corresponding probabilities.
\begin{definition}{(PDFA)}
A probabilistic DFA (PDFA) is a tuple $\mathcal{A}^{\mathbb{P}} = (\mathcal{A}, \delta_{\mathbb{P}}, F_{\mathbb{P}})$, where $\mathcal{A}$ is a DFA, $\delta_{\mathbb{P}} : \delta \rightarrow [0,1]$ assigns a probability to each transition in $\delta$, ensuring that $\sum_{\sigma \in \Sigma} \delta_{\mathbb{P}}(q, \sigma, \delta(q, \sigma)) = 1$ for every $q \in Q$. Additionally, $F_{\mathbb{P}} : Q \rightarrow [0,1]$ assigns a probability of terminating at each state, with $F_{\mathbb{P}}(q) = 0$ if $q \notin F$.
\end{definition}
Consider a word $\omega=\sigma_0\sigma_1...\sigma_{n-1}$ and its induced trace $\tau=q_0q_1...q_{n}$ on PDFA $\mathcal{A}^{\mathbb{P}}$. The probability of $\omega$ is given by
\begin{equation}
    P(\omega)=\prod^{n-1}_{i=0} \delta_{\mathbb{P}}(q_i,\sigma_i,q_{i+1}) \cdot F_{\mathbb{P}}(q_{n}).
\end{equation}
We say $\mathcal{A}^{\mathbb{P}}$ accepts $\omega$ iff $P(\omega) > 0$.
The language of $\mathcal{A}^{\mathbb{P}}$ is the set of words with non-zero probabilities.

\subsection{Problem}
Given a set of demonstrations $D=\{d_0,d_1,...,d_{m-1}\}$ we want to (1) learn a set of sub-goals $G$ representing the different steps of the demonstrated task; (2) learn a PDFA $\mathcal{A}^{\mathbb{P}}$ which represents the valid temporal orderings of sub-goals and the expert's preferences; (3) use the learned PDFA to compute a plan for an artificial agent to replicate the most preferred expert behavior.
We consider frequently observed partial states as sub-goals and use clustering to infer a set of sub-goals from demonstrations as detailed in Section \ref{subsec:sub-goal-inf}. In Section \ref{subsec:pdfa-inference}, we describe how a PDFA can be constructed. This PDFA can be used for planning as described in Section \ref{subsec:planning}.

\section{METHODOLOGY}
\label{sec:method}
\subsection{Sub-Goal Inference}
\label{subsec:sub-goal-inf}
Each sub-goal is defined by a subset of all features. We denote the candidate feature subsets as $\mathcal{C}(\mathcal{F})$, representing a collection of subsets of $F$. Each sub-goal is thus associated with one of these subsets: $\mathcal{F}_g \in \mathcal{C}(\mathcal{F})$ for all $g \in G$. We have access to a set of demonstrations $D = {d_0, d_1, \ldots, d_{m-1}}$. For each subset $\mathcal{F}_i \in \mathcal{C}(\mathcal{F})$, we construct a corresponding dataset by ensuring that, for each demonstration $d \in D$, each state $s \in d$ is transformed to a partial state $\hat{s} = \phi(\mathcal{F}_i, s)$ such that the partial state vector contains values only for the features $f \in \mathcal{F}_i$. This process results in a new collection of datasets $\hat{D} = {\hat{D}_0, \hat{D}1, \ldots, \hat{D}_{C-1}}$, where $C$ denotes the number of subsets in $\mathcal{C}(\mathcal{F})$.
We reason that genuine sub-goals should be frequently observed in the expert demonstrations. To identify these regions, we employ DBSCAN clustering \cite{ester1996density} on all datasets in $\hat{D}$. DBSCAN is robust to noise from incorrect object detections and does not require prior knowledge of the number of clusters, which is essential given that the number of sub-goals is unknown. After clustering, we eliminate clusters that intersect with initial states (i.e., states at the beginning of a demonstration). The remaining clusters form the set of sub-goals $G$. Note that each cluster, like a sub-goal, is defined by a point in a subspace of $\mathcal{F}$ and a radius.

\subsection{PDFA Inference}
\label{subsec:pdfa-inference}
First, we construct an alphabet $\Sigma$ such that each symbol $\sigma \in \Sigma$ represents a single sub-goal with $\sigma_g$ denoting the symbol corresponding to sub-goal $g$. Next, the set of demonstrations $D$ is converted into a set of words $\Omega$, such that each demonstration is represented by a sequence of sub-goals. The conversion process, detailed in Algorithm \ref{alg:traj_to_word}, initializes an empty word $\omega=\epsilon$. At each time step, we verify if any sub-goals are completed. Upon completion, the corresponding sub-goal symbol $\sigma_g$ is appended to $\omega$.
We assume that the preference for a particular word is correlated with its frequency of occurrence in $\Omega$. Therefore, preferences can be quantified as weights assigned to traces and normalized across the entire language, forming a probability distribution over $\mathcal{L}(\mathcal{A})$. A higher probability of an accepting word indicates greater preference by the demonstrator. Specifically, the demonstrator prefers $\omega$ over $\omega^{\prime}$ if and only if $P(\omega) > P(\omega^{\prime})$.
\\
\begin{algorithm}
\caption{Trajectory to word conversion}
\label{alg:traj_to_word}
\begin{algorithmic}
\REQUIRE sub-goal radius $r$, alphabet $\Sigma$, demonstration $d$
\STATE $\omega = \epsilon$
\FOR{$s \in d$}
\FOR{$g \in G$}
\IF{$\sigma_g \notin \omega$}
\STATE{$c, r = g$}
\IF{$||\phi(\mathcal{F}_g,s) - c||_2 \leq r$}
\STATE $\omega \leftarrow \omega \cdot \sigma_g$
\STATE break
\ENDIF
\ENDIF
\ENDFOR
\ENDFOR
\RETURN $\omega$
\end{algorithmic}
\end{algorithm}
\begin{algorithm}
\caption{DFA Inference}
\label{alg:dfa_infer}
\begin{algorithmic}
\REQUIRE alphabet $\Sigma$, set of words $\Omega$ representing expert demonstrations
\STATE $q_0 \leftarrow$ \verb|init_new_state()|
\STATE $Q \leftarrow \{ q_0 \}$
\STATE $F \leftarrow \emptyset$
\STATE $\delta(q, \sigma) \leftarrow \text{undefined} \;\;\;\; \forall q \in Q, \forall \sigma \in \Sigma$
\STATE $V \leftarrow \mathbf{0}_{2^{|\Sigma|} \times 2^{|\Sigma|}}$
\STATE $\psi(A) \leftarrow \text{undefined} \;\;\;\; \forall A \subseteq \Sigma$
\FOR{$\omega \in \Omega$}  
\STATE $\Lambda \leftarrow \emptyset$
\STATE $q \leftarrow q_0$
\FOR{$\sigma \in \omega$}
\STATE $\Lambda \leftarrow \Lambda \cup \{\sigma\}$
\IF{$\psi(\Lambda) \neq \text{undefined}$}
\IF{$\delta(q, \sigma) = \text{undefined}$}
\STATE $\delta(\hat{q}, \hat{\sigma}) \leftarrow \begin{cases}
      \psi(\Lambda) & \text{if} \; \hat{q} = q \text{ and } \hat{\sigma} = \sigma \\
      \delta(\hat{q}, \hat{\sigma}) & \text{otherwise}
    \end{cases}$
\ENDIF
\ELSE
\STATE $q_{\text{new}} \leftarrow$ \verb|init_new_state()|
\STATE $Q \leftarrow Q \cup \{q_{\text{new}}\}$
\STATE $\psi(A) \leftarrow \begin{cases}
    q_{\text{new}} & \text{if} \; A = \Lambda \\
    \psi(A) & \text{otherwise}
    \end{cases}$
\STATE $\delta(\hat{q}, \hat{\sigma}) \leftarrow \begin{cases}
      q_{\text{new}} & \text{if} \; \hat{q} = q \text{ and } \hat{\sigma} = \sigma \\
      \text{undefined} & \text{if} \; \hat{q} = q_{\text{new}} \\
      \delta(\hat{q}, \hat{\sigma}) & \text{otherwise}
    \end{cases}$
\ENDIF
\STATE $q \leftarrow \delta(q, \sigma)$
\STATE $V[q][q'] \leftarrow V[q][q'] + 1$
\ENDFOR
\IF{$q \notin F$}
\STATE $F \leftarrow F \cup \{q\}$
\ENDIF
\ENDFOR
\RETURN $(Q, q_0, \Sigma, \delta, F)$, $V$
\end{algorithmic}
\end{algorithm}
Algorithm \ref{alg:dfa_infer} describes the process for inferring a DFA that captures the sub-goal ordering observed in the demonstrations. Additionally, a frequency map $V$ is constructed to record the transition frequencies between each pair of DFA states.
First, the algorithm initializes a DFA $\mathcal{A}$ with the initial state $q_{0}$ and the alphabet $\Sigma$ representing the extracted sub-goals. 
The set of accepting states is initialized as an empty set $F = \emptyset$, and the transition function $\delta$ is initialized such that there are no valid transitions from the initial state $q_0$.
All values in the frequency map $V$ are initialized to zero.
A mapping function $\psi(A): 2^{\Sigma} \rightarrow Q$ is defined, which maps a subset of symbols to a DFA state, where $2^{\Sigma}$ denotes the power set of the alphabet $\Sigma$. This mapping $\psi$ is initially set to assign no subsets of $\Sigma$ to any states in $Q$.
\\
The algorithm then iterates over all words $\omega \in \Omega$. For each word, the current state $q$ is initialized to $q_0$.
An empty set $\Lambda$ is initialized to represent the set of already completed sub-goals.
For each symbol $\sigma$ in the word $\omega$, $\sigma$ is added to $\Lambda$. This process ensures that $\Lambda$ correctly represents the set of sub-goals completed so far.
Each word $\omega$ generates a trajectory that ends in a specific DFA state $q$.
When there are multiple valid sub-goal orderings, the algorithm ensures that all corresponding words generate trajectories ending in the same state, maintaining a direct relation between DFA states and completed sub-goals. If for the current state $q$ and symbol $\sigma$, $\psi(\Lambda) \neq \text{undefined}$ and $\delta(q, \sigma) = \text{undefined}$, it indicates that the same set of sub-goals has been completed in a different order. In such cases, only the transition function $\delta$ needs to be updated. If the current set of completed sub-goals $\Lambda$ has not been observed before, a new state $q_{\text{new}}$ should be added. The transition function $\delta$ is updated, and $\psi$ is modified to map $\Lambda$ to $q_{\text{new}}$. This algorithm assures that all demonstrations are represented by the DFA's language $\mathcal{L}(\mathcal{A})$.
\\
Each symbol $\sigma$ induces a transition from state $q$ to $q^{\prime}$, for each transition, the corresponding entry in $V$ is updated. The state reached at the end of each word $\omega \in \Omega$ is added to the set of accepting states $F$. If different demonstrations consist of different sets of completed sub-goals, $F$ will contain multiple accepting states. The probabilistic transition function $\delta_{\mathbb{P}}$ and the set of probabilistic accepting states $F_{\mathbb{P}}$ can be easily extracted from $V$, to construct a PDFA:
\begin{equation}
    \delta_{\mathbb{P}}(q,\sigma,q^{\prime}) = 
    \begin{cases}
      \dfrac{V[q][q^{\prime}]}{\sum_{q^{\prime\prime} \in Q} V[q][q^{\prime\prime}]} & \textrm{if} \; \delta(q,\sigma)=q^{\prime} \\
      0 & \text{otherwise},
    \end{cases}
\end{equation}
\begin{equation}
    F_{\mathbb{P}}(q) = 
    \begin{cases}
      \dfrac{\sum_{q^{\prime} \in Q} V[q^{\prime}][q]}{\sum_{q^{\prime} \in Q}\sum_{q^{\prime\prime} \in F} V[q^{\prime}][q^{\prime\prime}]} & \textrm{if} \; q \in F \\
      0 & \textrm{if} \; q \notin F.
    \end{cases}
\end{equation}
\begin{figure*}
    \centering
    \includegraphics[width=0.95\linewidth]{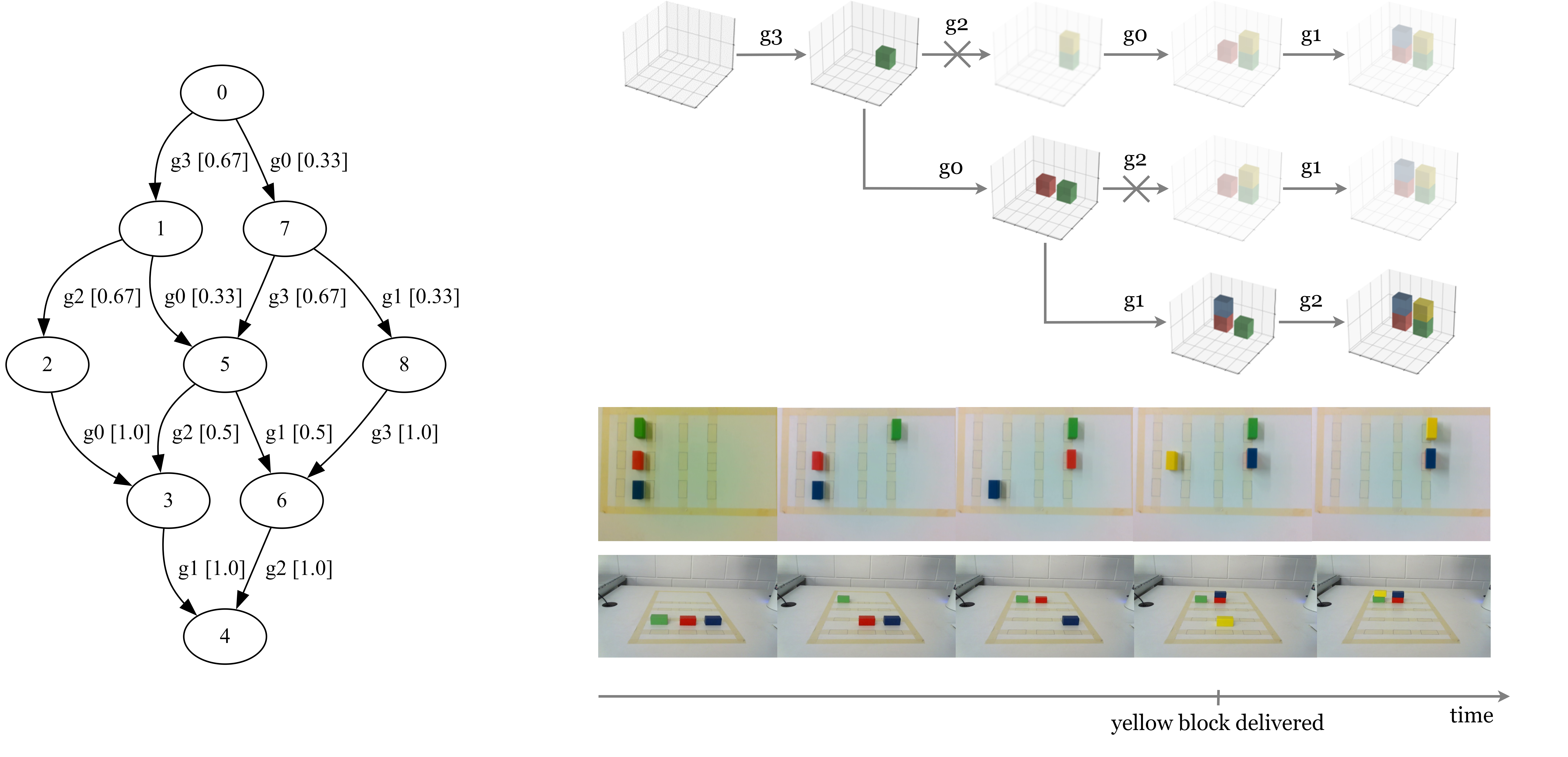}
    \vspace{-0.2cm}
    \caption{\textbf{Left}: Inferred PDFA (from 9 demonstrations) for a task involving the construction of two stacks of two blocks, with each block required to be placed in a designated location. \textbf{Top right}: The initial plan selects sub-goals based on the highest transition probabilities. However, during execution, the unavailability of the yellow block prevents the agent from completing sub-goal 2, necessitating two re-planning steps. \textbf{Bottom right}: Snapshots of the environment during task execution. The top row displays frames captured by the camera mounted on the robot's end effector, while the bottom row shows side-view frames captured by an additional camera.}
    \label{fig:planning_example}
\vspace{-0.6cm}
\end{figure*}

\subsection{Planning With PDFAs}
\label{subsec:planning}
Once a task is represented as a PDFA, our goal is to generate a plan that not only accomplishes the task but also aligns with the demonstrator’s preferences. We assume the existence of a low-level controller, such as a model-based controller or a trained RL policy, capable of achieving individual sub-goals. Each transition in the PDFA corresponds to a single execution of this controller, achieving the associated sub-goal. Consequently, replicating the expert's behavior is simplified to a process of greedily selecting the transition with the highest probability at each PDFA state, continuing this process until an accepting state is reached.

\section{EXPERIMENTS}
\label{sec:experiments}
We extensively evaluated our method on object manipulation tasks using unstructured human demonstrations. We deployed our approach on a physical robot to replicate expert behaviors and tested it in two simulated environments to demonstrate its versatility. The first simulated environment involved a surveillance task executed by a quadcopter using a path and motion planning algorithm for low-level control. The second involved using RL for low-level control with a two-jointed robot arm.

\subsection{Object Manipulation}
\begin{figure*}[t]
\begin{center}
    \subfigure[]{\includegraphics[width=0.15\textwidth]{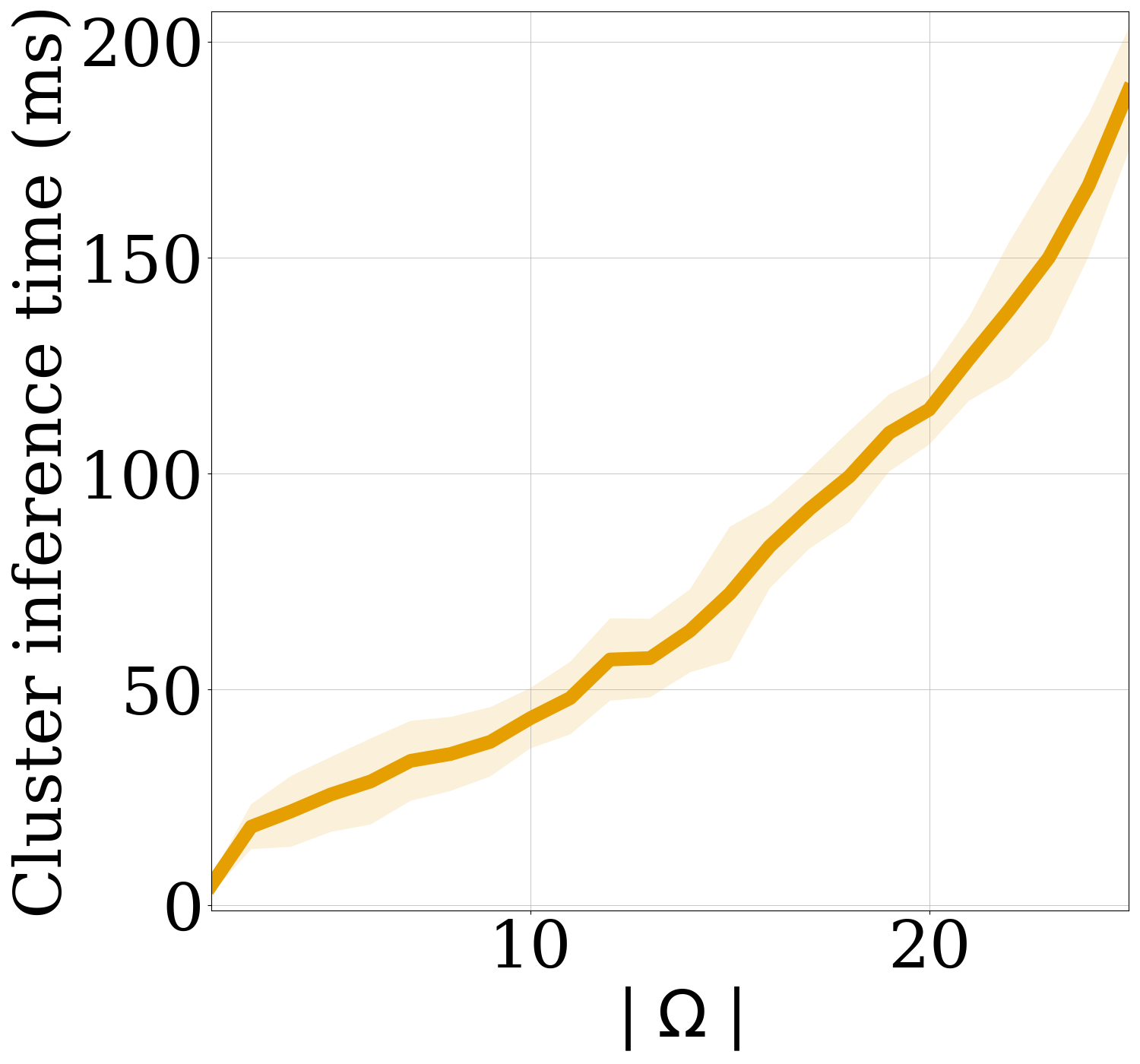}
    \label{fig:result_a}}
    \hspace{0.6cm}
    \subfigure[]{\includegraphics[width=0.15\textwidth]{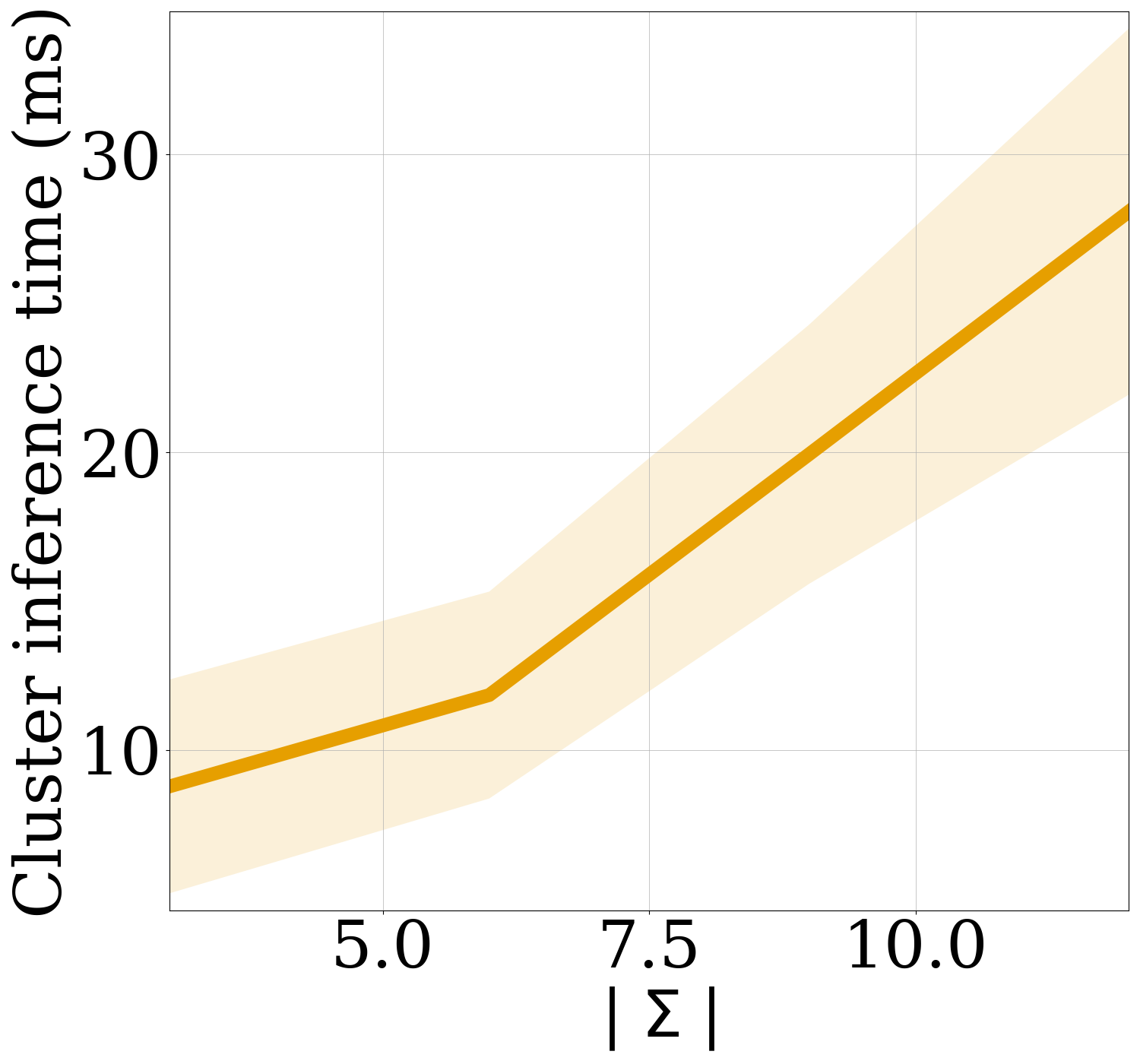}
    \label{fig:result_b}}
    \hspace{0.6cm}
    \subfigure[]{\includegraphics[width=0.15\textwidth]{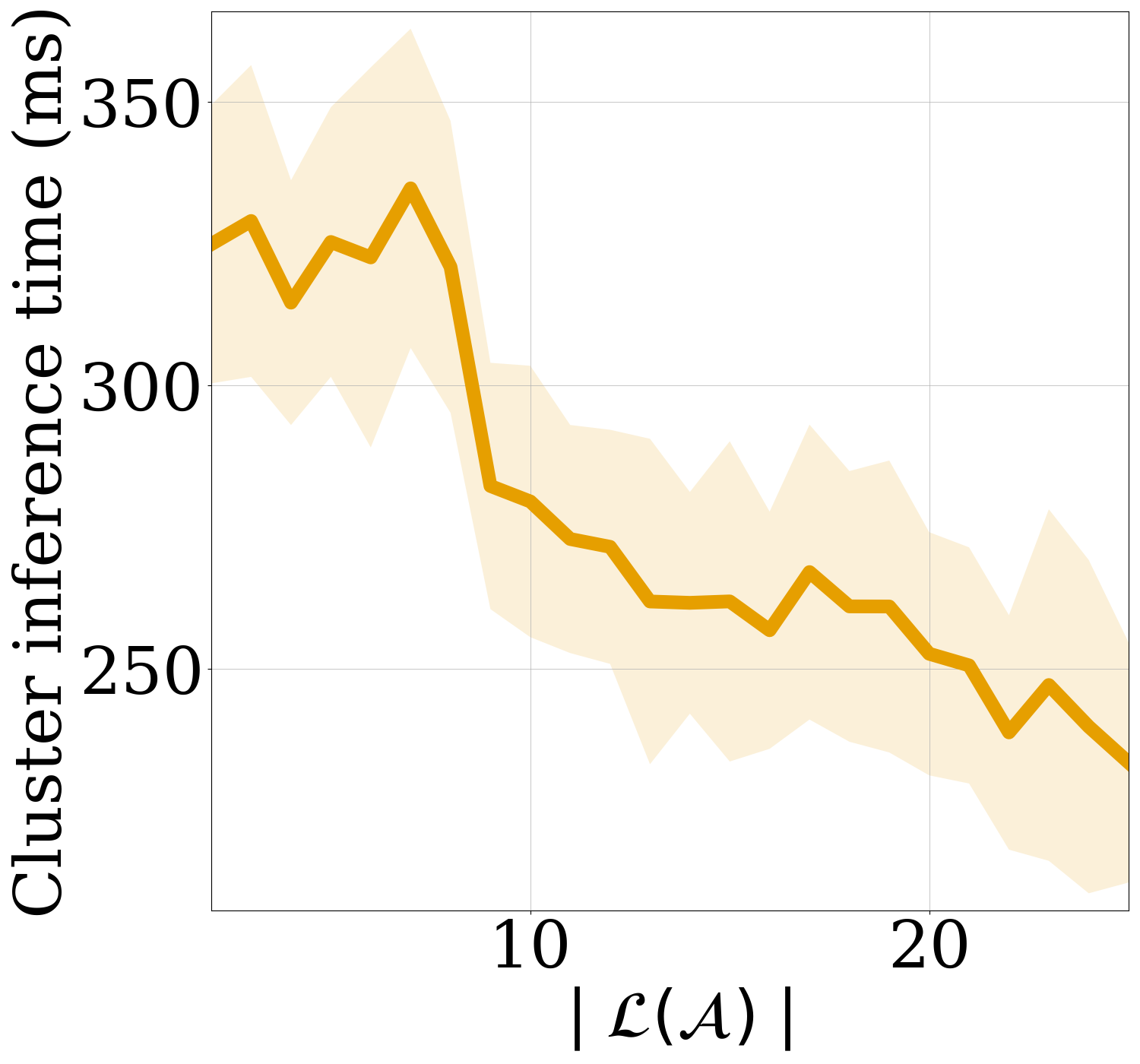}
    \label{fig:result_c}}
    \hspace{0.6cm}
    \subfigure[]{\includegraphics[width=0.15\textwidth]{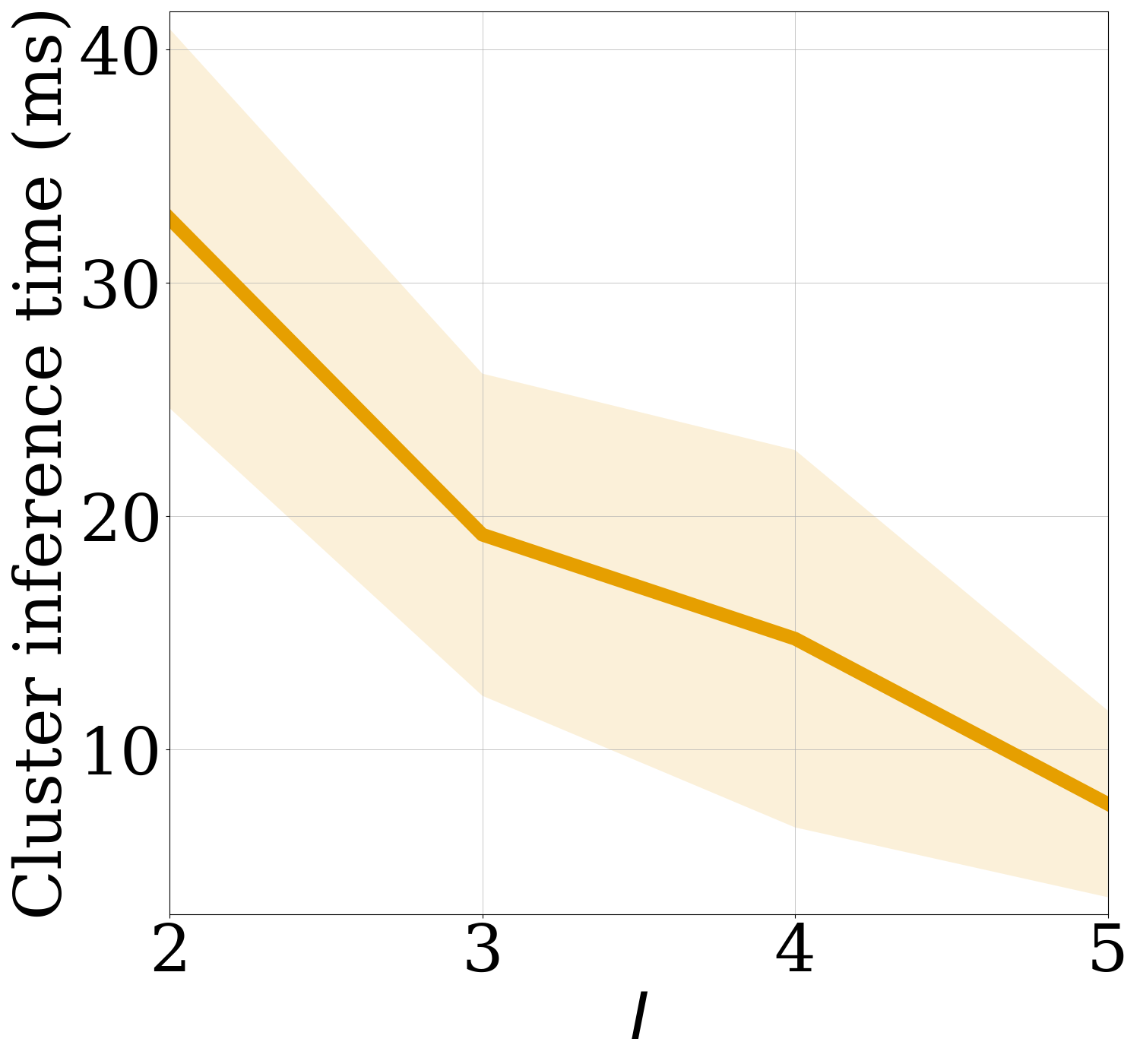}
    \label{fig:result_d}}\\
    \vspace{-0.2cm}
    \subfigure[]{\includegraphics[width=0.15\textwidth]{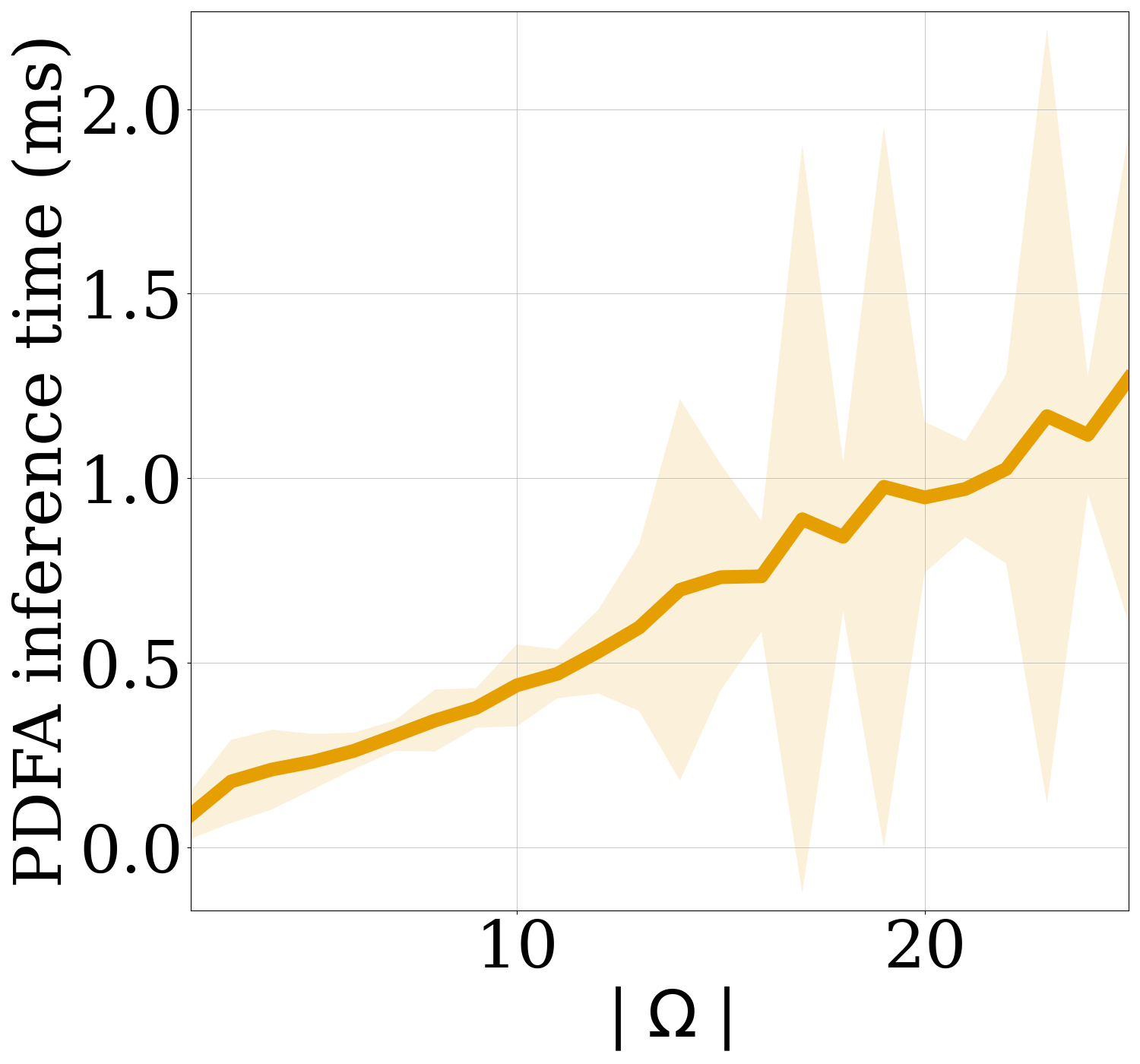}
    \label{fig:result_e}} 
    \hspace{0.6cm}
    \subfigure[]{\includegraphics[width=0.15\textwidth]{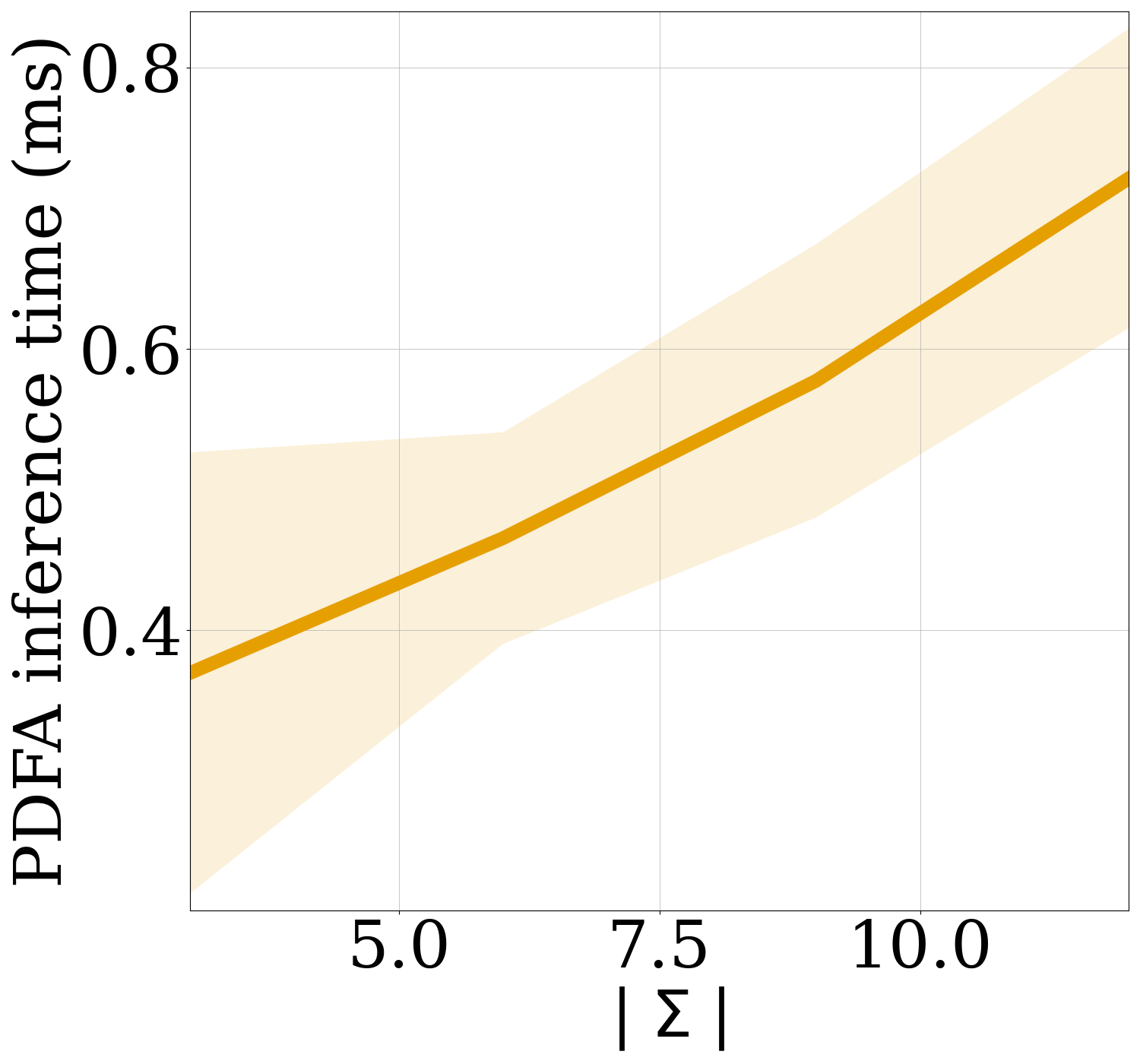}
    \label{fig:result_f}}
    \hspace{0.6cm}
    \subfigure[]{\includegraphics[width=0.15\textwidth]{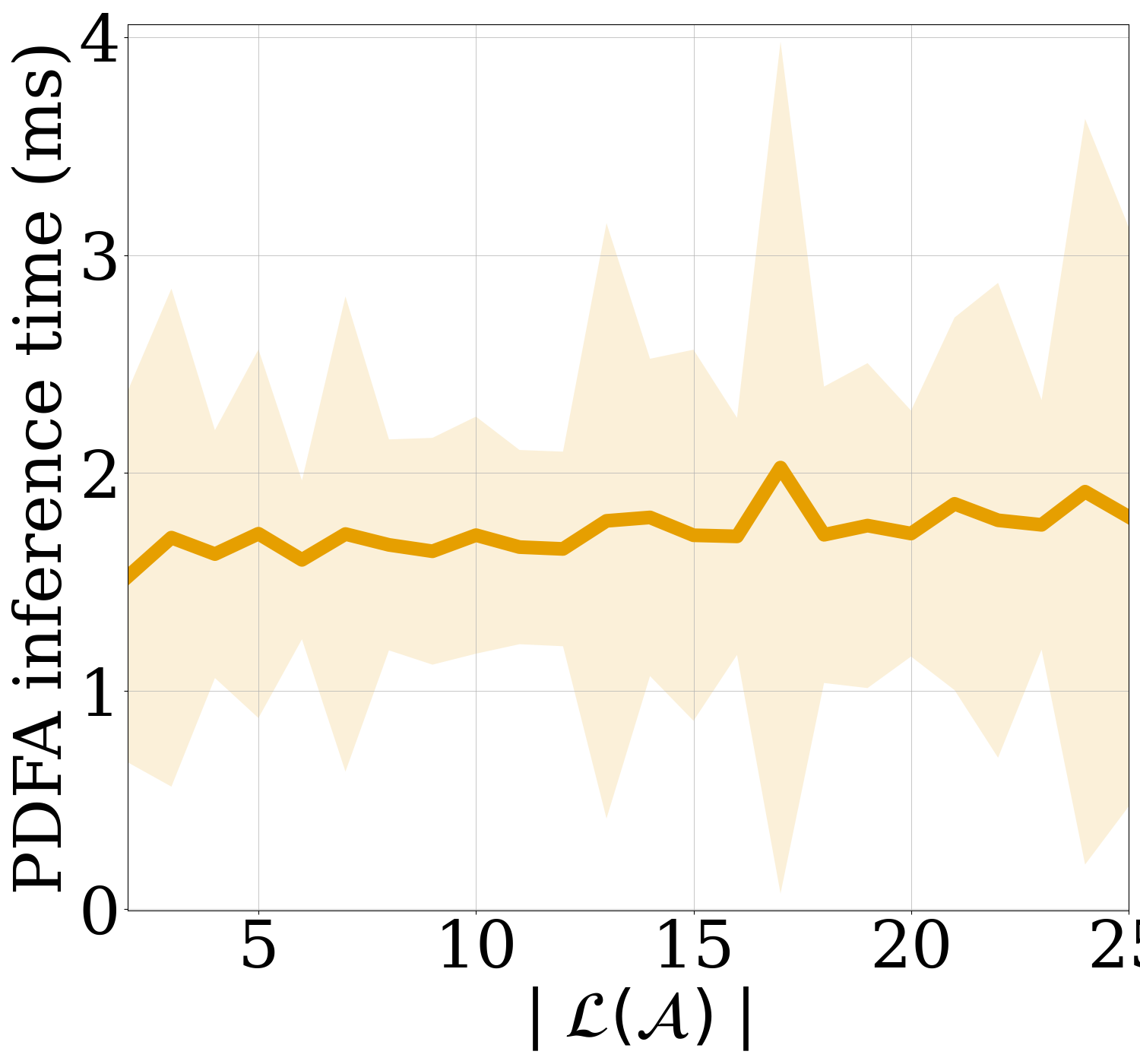}
    \label{fig:result_g}}
    \hspace{0.6cm}
    \subfigure[]{\includegraphics[width=0.15\textwidth]{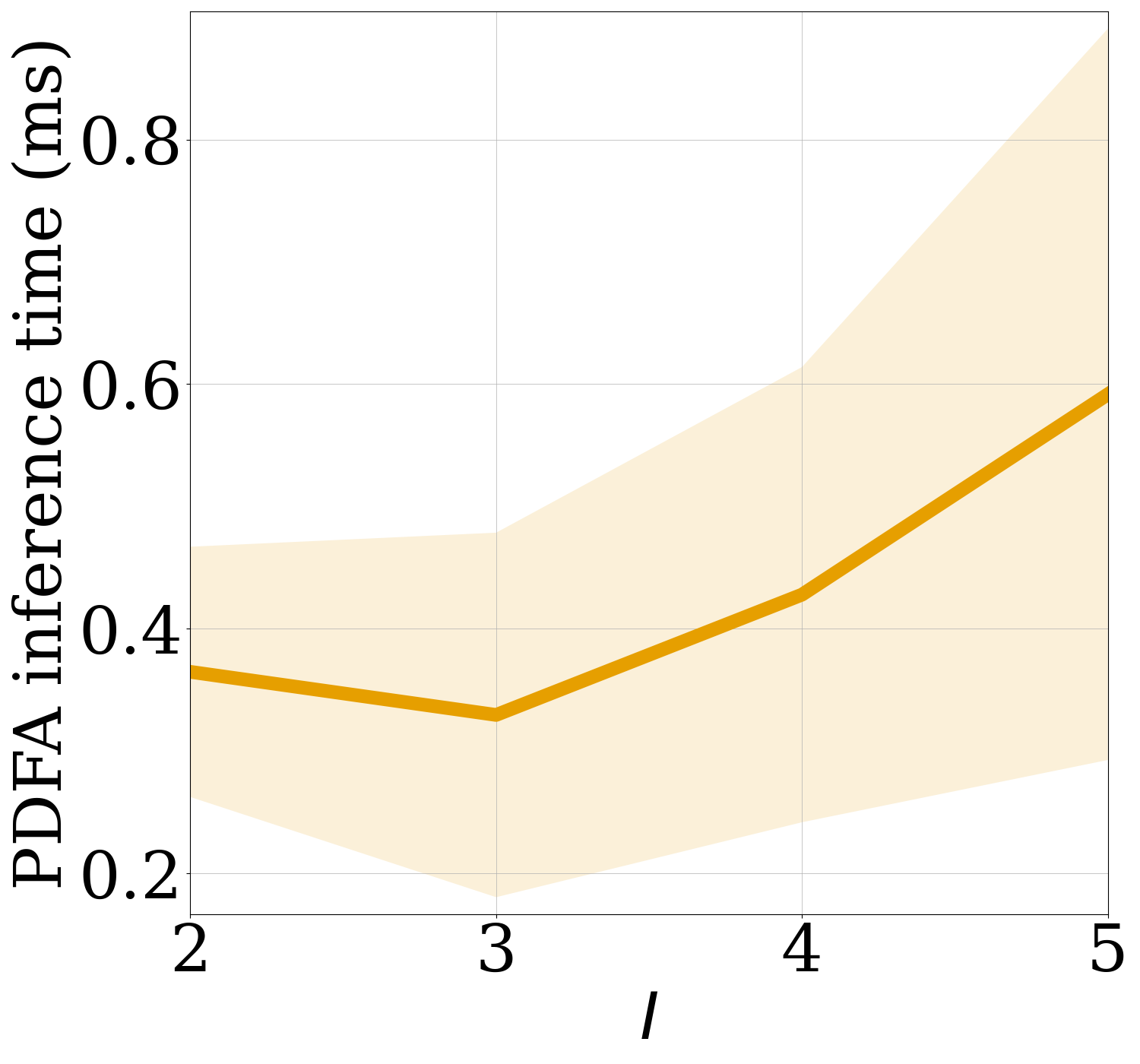}
    \label{fig:result_h}}
    \vspace{-0.2cm}
    \caption{Effect of different task properties on the cluster and PDFA inference time. The investigated properties are the number of demonstrations $\mid \Omega \mid$ (Fig. \ref{fig:result_a} and \ref{fig:result_e}), the number of sub-goals $\mid \Sigma \mid$ (Fig. \ref{fig:result_b} and \ref{fig:result_f}), the language size $\mathcal{L}(\mathcal{A})$ (Fig. \ref{fig:result_c} and \ref{fig:result_g}) and the number of objects $I$ (Fig. \ref{fig:result_d} and \ref{fig:result_h}).}
    \label{fig:results}
\end{center}
\vspace{-0.6cm}
\end{figure*}
We designed a series of multi-step object manipulation tasks using wooden blocks to evaluate our method. We utilized a Franka Emika Panda robot, with demonstrations performed by a human expert and recorded with an RGB-D camera mounted on the robot's end effector (the right picture in Fig. \ref{fig:method} depicts our setup). The end effector was positioned to observe the workspace from a predefined location during demonstrations. We used a thresholded depth image to detect the location and pose of objects, which, despite its simplicity, was effective for the simple shapes involved. RGB color detection was used to distinguish and track objects. The world state vector consists of the x-, y-, and z-coordinates of each object. 
For each object, a subset of features is added to the set of candidate subspaces $\mathcal{C}(\mathcal{F})$ that define its coordinates.
\\
For deployment, the robot's end effector was moved to the observation location to update the world state based on the current object positions. If an object was not detected, the corresponding features were set as undefined. After selecting a sub-goal, we used a low-level Cartesian impedance controller from the SERL Franka library \cite{luo2024serl} to complete the action. Post-action, the end effector returned to its initial position for another world state update. Because each sub-goal is defined as a location of a single object, we can assess the reachability of a sub-goal by checking if the corresponding object is detected.
\\
Fig. \ref{fig:planning_example} illustrates a learned PDFA for a task involving stacking two sets of blocks. The automaton indicates six valid orderings of the sub-goals, with the preferred sequence being g3, g2, g0, g1. The right side shows this initial plan in the top row. However, when it turns out that the next sub-goal is unreachable (in this case g2 because the yellow block is not detected), a new plan is generated (see second and third row).
\\
We investigated the impact of four task properties on clustering and PDFA inference time: the number of demonstrations $\mid \Omega \mid$, the number of sub-goals $\mid \Sigma \mid$, the language size of the extracted PDFA $\mid \mathcal{L}(\mathcal{A}) \mid$, and the number of objects $I$. Experiments were carefully designed to isolate the influence of each property while keeping others constant. Fig. \ref{fig:results} shows the results. To investigate the influence of the number of demonstrations $\mid \Omega \mid$, we used a task of placing four blocks at specified locations. For the clustering, inference time increases exponentially with the number of demonstrations. Our automata inference algorithm, however, shows linear time complexity due to the single loop over all demonstrations in Algorithm \ref{alg:dfa_infer}.
To assess the effect of the number of sub-goals $\mid \Sigma \mid$, we designed tasks involving the stacking and unstacking of three blocks. We conducted four experiments. For $\mid \Sigma \mid = 3$, building a stack of three blocks. For $\mid \Sigma \mid = 6$, first building the same stack and then a second stack using the same blocks but at a different location, with the top and bottom blocks swapped. Similarly, we designed experiments for $\mid \Sigma \mid = 9$ and $\mid \Sigma \mid = 12$. Both clustering and PDFA inference times scale linearly with $\mid \Sigma \mid$.
The language $\mathcal{L}(\mathcal{A})$ of an automaton $\mathcal{A}$ represents the variety of ways to perform a task. For this, we re-use the task of moving four blocks to specified locations. The demonstrations contain $4! = 24$ different ways of performing this task. As expected, the language size has no effect on PDFA inference time. Interestingly, clustering time decreases as $\mid \mathcal{L}(\mathcal{A}) \mid$ increases.
We also investigated the effect of the number of objects $I$. We expected no effect on both clustering and PDFA inference times. However, clustering time decreases with increasing $I$. This is due to our experiment design: for stacking multiple blocks (up to 5), to keep $\mid \Sigma \mid$ constant, we need to move blocks from the stack when $I < 5$. When $I = 5$, each feature subspace consists of a single sub-goal. For $I < 5$, some objects correspond to two or more sub-goals, making clustering harder and more time-consuming. There is a very small, unexpected increase in PDFA inference time.
\\
We compared our DFA inference procedure with a method using satisfiability (SAT) solvers \cite{heule2010exact} and another method for inferring LTL specifications from positive examples \cite{roy2023learning}. Our approach was significantly faster, with SAT solvers timing out on tasks with more than three sub-goals and the LTL method failing to infer correct specifications even for simple tasks. For both methods, we provide demonstrations already converted into words using the proposed sub-goal extraction method and algorithm \ref{alg:traj_to_word}.

\subsection{Drone Surveillance}
\begin{figure}
\begin{center}
    \subfigure[]{\includegraphics[width=0.15\textwidth]{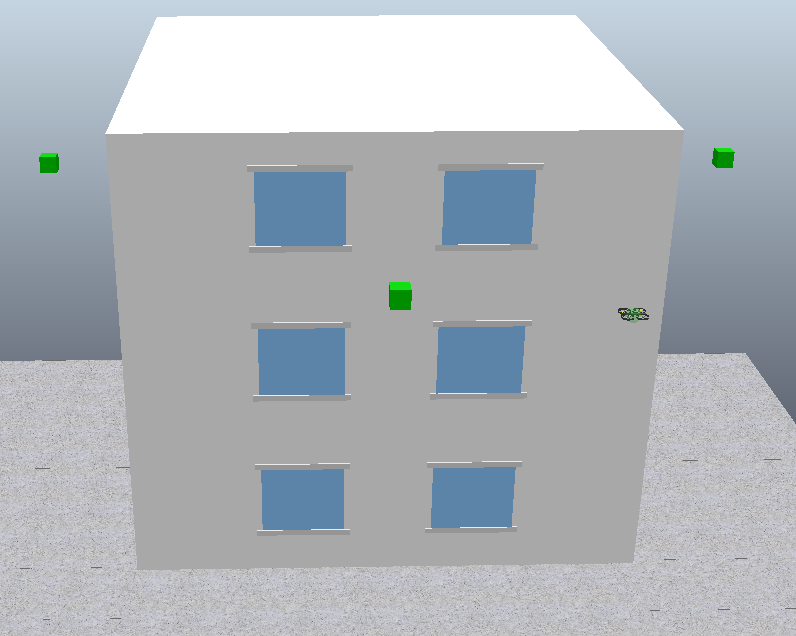}
    \label{fig:drone}}
    \hspace{0.6cm}
    \subfigure[]{\includegraphics[width=0.15\textwidth]{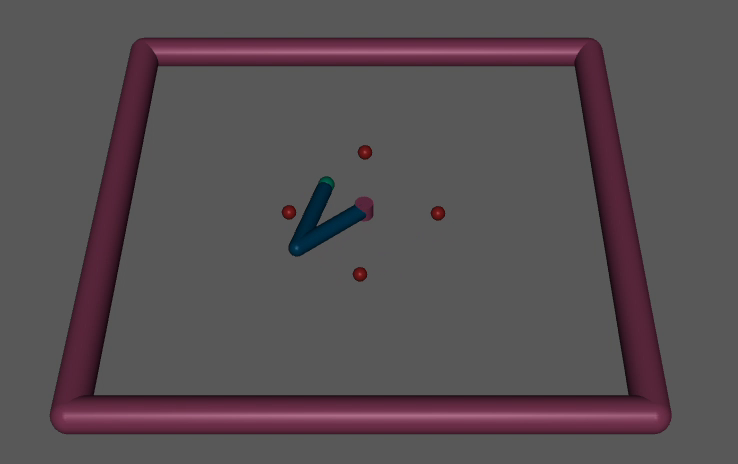}
    \label{fig:reacher}}
    \vspace{-0.2cm}
\end{center}
\vskip -0.1in
\caption{(a) drone surveillance (b) two-jointed robot arm.}
\vspace{-0.6cm}
\end{figure}

In a simulated environment inspired by \cite{chou2022learning}, we tasked a quadcopter with visiting three specified locations using a path and motion planning algorithm for low-level control. For this environment, a state is defined by the drone's coordinates and we defines $\mathcal{C}(\mathcal{F}) = \{\mathcal{F}\}$. Our method successfully identified the sub-goals and their order. The agent moved effectively between sub-goals using RTT motion planning and inverse kinematics. Fig. \ref{fig:drone} depicts this environment, the green boxes represent the inferred sub-goals.

\subsection{Two-Jointed Robot Arm}
To demonstrate compatibility with RL, we designed a task where a two-jointed robot arm's end effector moves to specific locations in a specified order. This environment is based on the Gymnasium Reacher environment \cite{towers2024gymnasium}, see Fig. \ref{fig:reacher} for a snapshot. We defined 4 sub-goals, one on the positive side of the x-axis, one on the negative side of the x-axis and similarly two on the y-axis. We defined a single feature subset consisting of the x- and y-coordinates of the end effector. We trained an RL agent on moving its end effector to random coordinates. Our method successfully extracted the four sub-goals and constructed a PDFA based on expert demonstrations, which alternated between clockwise and counterclockwise sub-goal sequences. Next, we successfully deployed the trained policy to perform the task.

\section{CONCLUSIONS}
\label{sec:conclustion}
In this paper, we presented a novel method for learning PDFAs from unstructured human demonstrations to capture task structure and demonstrator preferences allowing for robust replication of expert behaviors. Through extensive evaluations on object manipulation tasks with a physical robot arm, we demonstrated the effectiveness and efficiency of our method. Additionally, we validated the applicability of our approach in other domains by deploying it in simulated environments.
A limitation of our work is that we require domain knowledge to define the candidate feature subsets $\mathcal{C}(\mathcal{F})$ manually. Future work could investigate feature importance methods and dimensionality reduction techniques to automate this step. Instead of using a pre-trained a policy, a policy can be learned for each transition in the PDFA, or one for each sub-goal. This relates to hierarchical RL as each policy can be seen as an individual option \cite{konidaris2012robot}.
One could also condition the policy on the PDFA state similarly as \cite{icarte2022reward}.



\section*{ACKNOWLEDGMENT}
S.L. and P.S. acknowledge the financial support from the Flanders AI Research Program.


\bibliographystyle{IEEEtran}
\bibliography{ref}

\begin{thebibliography}{10}
\providecommand{\url}[1]{#1}
\csname url@rmstyle\endcsname
\providecommand{\newblock}{\relax}
\providecommand{\bibinfo}[2]{#2}
\providecommand\BIBentrySTDinterwordspacing{\spaceskip=0pt\relax}
\providecommand\BIBentryALTinterwordstretchfactor{4}
\providecommand\BIBentryALTinterwordspacing{\spaceskip=\fontdimen2\font plus
\BIBentryALTinterwordstretchfactor\fontdimen3\font minus \fontdimen4\font\relax}
\providecommand\BIBforeignlanguage[2]{{%
\expandafter\ifx\csname l@#1\endcsname\relax
\typeout{** WARNING: IEEEtran.bst: No hyphenation pattern has been}%
\typeout{** loaded for the language `#1'. Using the pattern for}%
\typeout{** the default language instead.}%
\else
\language=\csname l@#1\endcsname
\fi
#2}}

\bibitem{ho2016generative}
J.~Ho and S.~Ermon, ``Generative adversarial imitation learning,'' \emph{Advances in neural information processing systems}, vol.~29, 2016.

\bibitem{ghazanfari2020sequential}
B.~Ghazanfari, F.~Afghah, and M.~E. Taylor, ``Sequential association rule mining for autonomously extracting hierarchical task structures in reinforcement learning,'' \emph{IEEE Access}, vol.~8, pp. 11\,782--11\,799, 2020.

\bibitem{DBLP:conf/icra/ZhangMJLCGA18}
\BIBentryALTinterwordspacing
T.~Zhang, Z.~McCarthy, O.~Jow, D.~Lee, X.~Chen, K.~Goldberg, and P.~Abbeel, ``Deep imitation learning for complex manipulation tasks from virtual reality teleoperation,'' in \emph{2018 {IEEE} International Conference on Robotics and Automation, {ICRA} 2018, Brisbane, Australia, May 21-25, 2018}.\hskip 1em plus 0.5em minus 0.4em\relax {IEEE}, 2018, pp. 1--8. [Online]. Available: \url{https://doi.org/10.1109/ICRA.2018.8461249}
\BIBentrySTDinterwordspacing

\bibitem{kober2009learning}
J.~Kober and J.~Peters, ``Learning motor primitives for robotics,'' in \emph{2009 IEEE International Conference on Robotics and Automation}.\hskip 1em plus 0.5em minus 0.4em\relax IEEE, 2009, pp. 2112--2118.

\bibitem{chou2022learning}
G.~Chou, N.~Ozay, and D.~Berenson, ``Learning temporal logic formulas from suboptimal demonstrations: theory and experiments,'' \emph{Autonomous Robots}, vol.~46, no.~1, pp. 149--174, 2022.

\bibitem{liu2024interpretable}
\BIBentryALTinterwordspacing
W.~Liu, D.~Li, E.~Aasi, R.~Tron, and C.~Belta, ``Interpretable generative adversarial imitation learning,'' \emph{CoRR}, vol. abs/2402.10310, 2024. [Online]. Available: \url{https://doi.org/10.48550/arXiv.2402.10310}
\BIBentrySTDinterwordspacing

\bibitem{jha2019telex}
S.~Jha, A.~Tiwari, S.~A. Seshia, T.~Sahai, and N.~Shankar, ``Telex: learning signal temporal logic from positive examples using tightness metric,'' \emph{Formal Methods in System Design}, vol.~54, pp. 364--387, 2019.

\bibitem{shah2018bayesian}
A.~Shah, P.~Kamath, J.~A. Shah, and S.~Li, ``Bayesian inference of temporal task specifications from demonstrations,'' \emph{Advances in Neural Information Processing Systems}, vol.~31, 2018.

\bibitem{roy2023learning}
R.~Roy, J.-R. Gaglione, N.~Baharisangari, D.~Neider, Z.~Xu, and U.~Topcu, ``Learning interpretable temporal properties from positive examples only,'' in \emph{Proceedings of the AAAI Conference on Artificial Intelligence}, vol.~37, no.~5, 2023, pp. 6507--6515.

\bibitem{baier2008principles}
C.~Baier and J.-P. Katoen, \emph{Principles of model checking}.\hskip 1em plus 0.5em minus 0.4em\relax MIT press, 2008.

\bibitem{lang1998results}
K.~J. Lang, B.~A. Pearlmutter, and R.~A. Price, ``Results of the abbadingo one dfa learning competition and a new evidence-driven state merging algorithm,'' in \emph{International Colloquium on Grammatical Inference}.\hskip 1em plus 0.5em minus 0.4em\relax Springer, 1998, pp. 1--12.

\bibitem{petasis2004eg}
G.~Petasis, G.~Paliouras, C.~D. Spyropoulos, and C.~Halatsis, ``eg-grids: Context-free grammatical inference from positive examples using genetic search,'' in \emph{Grammatical Inference: Algorithms and Applications: 7th International Colloquium, ICGI 2004, Athens, Greece, October 11-13, 2004. Proceedings 7}.\hskip 1em plus 0.5em minus 0.4em\relax Springer, 2004, pp. 223--234.

\bibitem{watanabe2021probabilistic}
K.~Watanabe, N.~Renninger, S.~Sankaranarayanan, and M.~Lahijanian, ``Probabilistic specification learning for planning with safety constraints,'' in \emph{2021 IEEE/RSJ International Conference on Intelligent Robots and Systems (IROS)}.\hskip 1em plus 0.5em minus 0.4em\relax IEEE, 2021, pp. 6558--6565.

\bibitem{DBLP:conf/rss/MandlekarXMS020}
\BIBentryALTinterwordspacing
A.~Mandlekar, D.~Xu, R.~Mart{\'{\i}}n{-}Mart{\'{\i}}n, S.~Savarese, and L.~Fei{-}Fei, ``{GTI:} learning to generalize across long-horizon tasks from human demonstrations,'' in \emph{Robotics: Science and Systems XVI, Virtual Event / Corvalis, Oregon, USA, July 12-16, 2020}, M.~Toussaint, A.~Bicchi, and T.~Hermans, Eds., 2020. [Online]. Available: \url{https://doi.org/10.15607/RSS.2020.XVI.061}
\BIBentrySTDinterwordspacing

\bibitem{schaal2006dynamic}
S.~Schaal, ``Dynamic movement primitives-a framework for motor control in humans and humanoid robotics,'' in \emph{Adaptive motion of animals and machines}.\hskip 1em plus 0.5em minus 0.4em\relax Springer, 2006, pp. 261--280.

\bibitem{abbeel2004apprenticeship}
P.~Abbeel and A.~Y. Ng, ``Apprenticeship learning via inverse reinforcement learning,'' in \emph{Proceedings of the twenty-first international conference on Machine learning}, 2004, p.~1.

\bibitem{baert2023maximum}
M.~Baert, P.~Mazzaglia, S.~Leroux, and P.~Simoens, ``Maximum causal entropy inverse constrained reinforcement learning,'' \emph{arXiv preprint arXiv:2305.02857}, 2023.

\bibitem{arnold2017value}
T.~Arnold and D.~Kasenberg, ``Value alignment or misalignment--what will keep systems accountable?'' in \emph{AAAI Workshop on AI, Ethics, and Society}, 2017.

\bibitem{pnueli1977temporal}
A.~Pnueli, ``The temporal logic of programs,'' in \emph{18th annual symposium on foundations of computer science (sfcs 1977)}.\hskip 1em plus 0.5em minus 0.4em\relax ieee, 1977, pp. 46--57.

\bibitem{kim2017collaborative}
J.~Kim, C.~Banks, and J.~Shah, ``Collaborative planning with encoding of users' high-level strategies,'' in \emph{Proceedings of the AAAI Conference on Artificial Intelligence}, vol.~31, no.~1, 2017.

\bibitem{kress2009temporal}
H.~Kress-Gazit, G.~E. Fainekos, and G.~J. Pappas, ``Temporal-logic-based reactive mission and motion planning,'' \emph{IEEE transactions on robotics}, vol.~25, no.~6, pp. 1370--1381, 2009.

\bibitem{aksaray2016q}
D.~Aksaray, A.~Jones, Z.~Kong, M.~Schwager, and C.~Belta, ``Q-learning for robust satisfaction of signal temporal logic specifications,'' in \emph{2016 IEEE 55th Conference on Decision and Control (CDC)}.\hskip 1em plus 0.5em minus 0.4em\relax IEEE, 2016, pp. 6565--6570.

\bibitem{li2017reinforcement}
X.~Li, C.-I. Vasile, and C.~Belta, ``Reinforcement learning with temporal logic rewards,'' in \emph{2017 IEEE/RSJ International Conference on Intelligent Robots and Systems (IROS)}.\hskip 1em plus 0.5em minus 0.4em\relax IEEE, 2017, pp. 3834--3839.

\bibitem{xiong2022constrained}
Z.~Xiong, J.~Eappen, A.~H. Qureshi, and S.~Jagannathan, ``Constrained hierarchical deep reinforcement learning with differentiable formal specifications,'' 2022.

\bibitem{bombara2021offline}
G.~Bombara and C.~Belta, ``Offline and online learning of signal temporal logic formulae using decision trees,'' \emph{ACM Transactions on Cyber-Physical Systems}, vol.~5, no.~3, pp. 1--23, 2021.

\bibitem{kong2016temporal}
Z.~Kong, A.~Jones, and C.~Belta, ``Temporal logics for learning and detection of anomalous behavior,'' \emph{IEEE Transactions on Automatic Control}, vol.~62, no.~3, pp. 1210--1222, 2016.

\bibitem{yan2022neuro}
R.~Yan, T.~Ma, A.~Fokoue, M.~Chang, and A.~Julius, ``Neuro-symbolic models for interpretable time series classification using temporal logic description,'' in \emph{2022 IEEE International Conference on Data Mining (ICDM)}.\hskip 1em plus 0.5em minus 0.4em\relax IEEE, 2022, pp. 618--627.

\bibitem{vazquez2018learning}
M.~Vazquez-Chanlatte, S.~Jha, A.~Tiwari, M.~K. Ho, and S.~Seshia, ``Learning task specifications from demonstrations,'' \emph{Advances in neural information processing systems}, vol.~31, 2018.

\bibitem{chiu2023temporal}
T.-Y. Chiu, J.~Le~Ny, and J.-P. David, ``Temporal logic explanations for dynamic decision systems using anchors and monte carlo tree search,'' \emph{Artificial Intelligence}, vol. 318, p. 103897, 2023.

\bibitem{araki2019learning}
B.~Araki, K.~Vodrahalli, T.~Leech, C.-I. Vasile, M.~D. Donahue, and D.~L. Rus, ``Learning to plan with logical automata,'' 2019.

\bibitem{ekvall2008robot}
S.~Ekvall and D.~Kragic, ``Robot learning from demonstration: a task-level planning approach,'' \emph{International Journal of Advanced Robotic Systems}, vol.~5, no.~3, p.~33, 2008.

\bibitem{niekum2015learning}
S.~Niekum, S.~Osentoski, G.~Konidaris, S.~Chitta, B.~Marthi, and A.~G. Barto, ``Learning grounded finite-state representations from unstructured demonstrations,'' \emph{The International Journal of Robotics Research}, vol.~34, no.~2, pp. 131--157, 2015.

\bibitem{konidaris2012robot}
G.~Konidaris, S.~Kuindersma, R.~Grupen, and A.~Barto, ``Robot learning from demonstration by constructing skill trees,'' \emph{The International Journal of Robotics Research}, vol.~31, no.~3, pp. 360--375, 2012.

\bibitem{grollman2010incremental}
D.~H. Grollman and O.~C. Jenkins, ``Incremental learning of subtasks from unsegmented demonstration,'' in \emph{2010 IEEE/RSJ International Conference on Intelligent Robots and Systems}.\hskip 1em plus 0.5em minus 0.4em\relax IEEE, 2010, pp. 261--266.

\bibitem{pirk2020modeling}
\BIBentryALTinterwordspacing
S.~Pirk, K.~Hausman, A.~Toshev, and M.~Khansari, ``Modeling long-horizon tasks as sequential interaction landscapes,'' in \emph{4th Conference on Robot Learning, CoRL 2020, 16-18 November 2020, Virtual Event / Cambridge, MA, {USA}}, ser. Proceedings of Machine Learning Research, J.~Kober, F.~Ramos, and C.~J. Tomlin, Eds., vol. 155.\hskip 1em plus 0.5em minus 0.4em\relax {PMLR}, 2020, pp. 471--484. [Online]. Available: \url{https://proceedings.mlr.press/v155/pirk21a.html}
\BIBentrySTDinterwordspacing

\bibitem{manschitz2014learning}
S.~Manschitz, J.~Kober, M.~Gienger, and J.~Peters, ``Learning to sequence movement primitives from demonstrations,'' in \emph{2014 IEEE/RSJ International Conference on Intelligent Robots and Systems}.\hskip 1em plus 0.5em minus 0.4em\relax IEEE, 2014, pp. 4414--4421.

\bibitem{mohseni2019simultaneous}
A.~Mohseni-Kabir, C.~Li, V.~Wu, D.~Miller, B.~Hylak, S.~Chernova, D.~Berenson, C.~Sidner, and C.~Rich, ``Simultaneous learning of hierarchy and primitives for complex robot tasks,'' \emph{Autonomous Robots}, vol.~43, pp. 859--874, 2019.

\bibitem{lang1992random}
K.~J. Lang, ``Random dfa's can be approximately learned from sparse uniform examples,'' in \emph{Proceedings of the fifth annual workshop on Computational learning theory}, 1992, pp. 45--52.

\bibitem{rivest1989inference}
R.~L. Rivest and R.~E. Schapire, ``Inference of finite automata using homing sequences,'' in \emph{Proceedings of the twenty-first annual ACM symposium on Theory of computing}, 1989, pp. 411--420.

\bibitem{dupont1994regular}
P.~Dupont, ``Regular grammatical inference from positive and negative samples by genetic search: the gig method,'' in \emph{International Colloquium on Grammatical Inference}.\hskip 1em plus 0.5em minus 0.4em\relax Springer, 1994, pp. 236--245.

\bibitem{ester1996density}
M.~Ester, H.-P. Kriegel, J.~Sander, X.~Xu, \emph{et~al.}, ``A density-based algorithm for discovering clusters in large spatial databases with noise,'' in \emph{kdd}, 1996, pp. 226--231.

\bibitem{luo2024serl}
J.~Luo, Z.~Hu, C.~Xu, Y.~L. Tan, J.~Berg, A.~Sharma, S.~Schaal, C.~Finn, A.~Gupta, and S.~Levine, ``Serl: A software suite for sample-efficient robotic reinforcement learning,'' 2024.

\bibitem{heule2010exact}
M.~J. Heule and S.~Verwer, ``Exact dfa identification using sat solvers,'' in \emph{Grammatical Inference: Theoretical Results and Applications: 10th International Colloquium, ICGI 2010, Valencia, Spain, September 13-16, 2010. Proceedings 10}.\hskip 1em plus 0.5em minus 0.4em\relax Springer, 2010, pp. 66--79.

\bibitem{towers2024gymnasium}
M.~Towers, A.~Kwiatkowski, J.~Terry, J.~U. Balis, G.~De~Cola, T.~Deleu, M.~Goul{\~a}o, A.~Kallinteris, M.~Krimmel, A.~KG, \emph{et~al.}, ``Gymnasium: A standard interface for reinforcement learning environments,'' \emph{arXiv preprint arXiv:2407.17032}, 2024.

\bibitem{icarte2022reward}
R.~T. Icarte, T.~Q. Klassen, R.~Valenzano, and S.~A. McIlraith, ``Reward machines: Exploiting reward function structure in reinforcement learning,'' \emph{Journal of Artificial Intelligence Research}, vol.~73, pp. 173--208, 2022.

\end{thebibliography}

\end{document}